# Bridging the gap between Legal Practitioners and Knowledge Engineers using semi-formal KR

Shashishekar Ramakrishna and Adrian Paschke

Department of Computer Science Freie Universitaet Berlin,
Königin-Luise-Straße 24/26,
14195 Berlin, Germany
shashi792@gmail.com, paschke@inf.fu-berlin.de

**Abstract.** The use of Structured English as a computation independent knowledge representation format for non-technical users in business rules representation has been proposed in OMG's Semantics and Business Vocabulary Representation (SBVR). In the legal domain we face a similar problem. Formal representation languages, such as OASIS' LegalRuleML and legal ontologies (LKIF, legal OWL2 ontologies etc.) support the technical knowledge engineer and the automated reasoning. But, they can be hardly used directly by the legal domain experts who do not have a computer science background. In this paper we adapt the SBVR Structured English approach for the legal domain and implement a proof-of-concept, called KR4IPLaw, which enables legal domain experts to represent their knowledge in Structured English in a computational independent and hence, for them, more usable way. The benefit of this approach is that the underlying pre-defined semantics of the Structured English approach makes transformations into formal languages such as OASIS LegalRuleML and OWL2 ontologies possible. We exemplify our approach in the domain of patent law.

**Keywords:** SBVR, Structured English, Legal Norms, LegalRuleML

## 1 Introduction

There exists a gap concerning the understanding of the knowledge from a particular domain between a domain expert and a knowledge engineer, who models such domain knowledge – often in a structured, formal language - for its use in (semi-/) automated reasoning. Also, such knowledge representations modeled by the knowledge engineer are not generally automatically reusable outside the specific context for which the knowledge representation was originally developed. Such a problem can be easily seen in legal domain, wherein, the cost associated with not reducing such gaps is substantially high.

This paper contributes with an approach using a Structured English knowledge representation language with which this gap can be substantially minimized. The paper also introduces a tool, called KR4IPLaw, which is intended as a proof-of-concept implementation for the proposed approach. To illustrate it, in this paper, we concentrate on one branch of law dealing with technical innovations, namely patent law, esp. only a subset of laws used by an examiner for evaluating a patent application.

The paper is organized as follows: Section 2 describes the existing state of the art. Section 3 describes how the existing Structured English approach could be adapted and applied in the context of the legal domain. In section 4, we introduce the proof-of-concept tool KR4IPLaw. By a concrete use case example we show how procedural patent rules can be semi-formally represented using Structured English.

## 2 Related Work

The OMG's Model Driven Architecture, 'MDA' [1] provides a basis for representing information on different layers of knowledge representation models ( CIM, PIM and PSM).

Semantic Business Vocabulary and Business Rules, 'SBVR' [2], is an ISO terminological dictionary (vocabulary) for defining business concepts and rules. SBVR works on the Computational Independent Model layer of the OMG's MDA. It supports the use of Structured English (SE), a computational-independent English (natural) language having the syntax of a structured programming for representing business vocabularies and business rules. SBVR captures the structural and behavioral aspects of business processes, as well as the policies that should guide the agents' behavior in certain situations. A core idea of business rules formally supported by SBVR is the following: Rules build on facts, and facts build on concepts as expressed by terms. Terms express business concepts; facts make assertions about these concepts; rules constrain and support these facts [2]. Fig 1 depicts the relation of SBVR and OMG's MDA.

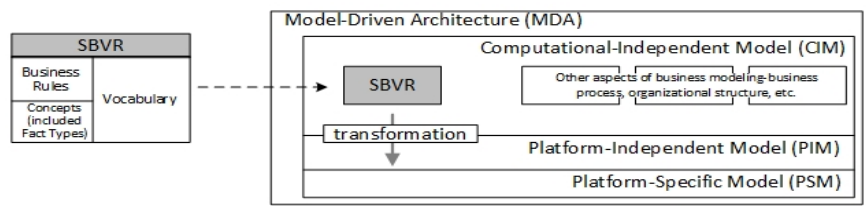

**Fig.1.** SBVR position in MDA (adapted from [2]).

The power of SBVR is disclosed by the fact that the SBVR specification itself was formally written in SBVR Structured English, 'SSE'[1]. The use of SBVR in legal domain was proposed by Johnsen and Berre in [3, 4]. In [5] we showed how OMG's MDA could be viewed in the domain of patent law, wherein, we provided the first ideas on using SBVR –SE in patent law domain.

## 3 Semi-formal KR in legal domain

We adapt the approach of the OMG Semantic Business Vocabulary and Business Rules [2] (OMG SBVR) standard to the patent law domain. Fig 2 shows the overview of such an adoption.

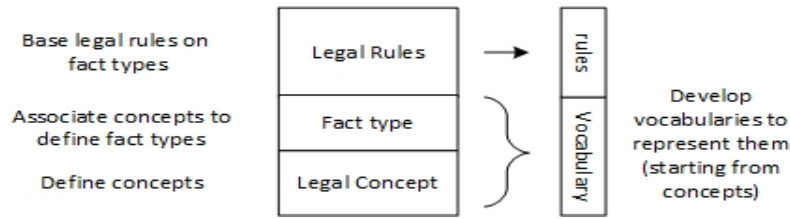

**Fig. 2.** Building legal vocabulary.

SBVR defines the vocabulary and rules for describing the legal semantics using SSE. Even though SSE does not provide all the expressivity required for translating the procedural rules into a formal reasoning, the simple approach of SSE helps the end users (i.e. the domain experts and legal practitioners) to define their legal vocabularies & rules in a more understandable manner, which at the same time can be also interpretable by the computer.

Like in SBVR, we define the legal (procedural/substantive) rules in a structured natural language (a Structured English syntax) using predefined **legal vocabularies,** consisting of legal concepts (concepts which have a meaning in the legal tradition, e.g. claim construction vocabulary) in template-based **legal rules**.

"Legal vocabulary" and "legal (argumentation) rules" are made of:

- **Noun concepts**, which correspond to legal concepts
- **Verb concepts**, which correspond to relationships between legal concepts
- **Definitional rules**, which constrain these relationships so that they can be used to define consistent and complete arguments

Legal concepts represented by noun concepts must be explicitly defined with the intended semantics given in an authoritative source or otherwise acknowledge by implicit pragmatic understanding (the ordinary natural language meaning of the term used). Verb concepts can only use such recognized noun concepts as their terms.

The legal rules can then be constructed using the "*if ... then ...*", "*at least*", "*each*" as well as definitional alethic and behavioral deontic legal norm modalities ("*obliged*", "*permitted*" …), etc. The following example in the next section illustrates its use.

## 4   KR4IPLaw

KR4IPLaw (Knowledge Representation for Intellectual Property Law) is a tool implemented on the Eclipse 4.3 (Kepler) platform. It currently supports SBVR 1.0 metamodel and is built based on SBeaVeR [6], SWeDE and OntoSphere3D [7]. The long term goal of this tool is to act as an interface, which can be easily handled by legal practitioners and is capable enough to provide all the necessary inputs for an knowledge engineer to model legal rules for (semi-/) automated reasoning after transforming them into a Platform Independent Model (PIM) rules. Such a tool will accommodate all the possible KR's (from formal to natural language) and act as a bridge between the legal practitioner and knowledge modeler.

To illustrate the use of SSE in the legal domain, we consider the legal (procedural) rules followed by an examiner in evaluating the essential subject matter requirement as defined in Paragraph ¶ 7.33.01 of United States Patent Law [8]- which states as follow:

> *¶ 7.33.01 Rejection, 35 U.S.C. 112, 1st Paragraph, Essential Subject Matter Missing From Claims (Enablement)*
>
> *Claim [1] rejected under 35 U.S.C. 112, first paragraph, as based on a disclosure which is not enabling. [2] critical or essential to the practice of the invention, but not included in the claim(s) is not enabled by the disclosure.*
> *1. This rejection must be preceded by form paragraph 7.30.01 or 7.103.*
> *2. In bracket 2, recite the subject matter omitted from the claims.*

Using SBVR Structured English

*Legal Concepts (Noun concepts defined in green and individual noun concepts are defined in dark-green starting with Capital letters)*

| claim | |
|---|---|
| Definition | Define the invention and are what aspects are legally enforceable |
| Dictionary basis | patentlaw |
| Source | based on USPTOGlossary |
| General_concept | patent |

Building on the same lines, we obtain other legal concepts like:

examiner, office_action, paragraphs, statement, argument, date, drawing, applicant, effective_feature, invention essential_subject_matter_requirement,

Paragarph_7_33_01

*Legal Facts (verb concepts are defined in blue)*

office_action *includes* paragraphs
claim *is_rejected_under* essential_subject_matter_requirement
office_action *include* statement
applicant *conceals* effective_feature
effective_feature *is_about* the invention
examiner *applies* Paragarph_7_33_01
examiner *rejects* the claim

> *Legal (procedural) rules (for ¶ 7.33.01):*
>
> 1. It is obligatory that examiner *rejects* the claim and office_action *includes* paragraphs Paragarph_7_33_01 if claim *is_rejected_under* essential_subject_matter_requirement
>
> 2. It is obligatory that office_action *include* statement and argument and date and drawing if claim *is_rejected_under* Paragarph_7_33_01
>
> 3. It is necessary that examiner *applies* Paragarph_7_33_01 if applicant *conceals* effective_feature and effective_feature *is_about* the invention

Fig 3 gives an overview of the KR4IPLaw tool from a legal practitioner's perspective, wherein, legal practitioner/domain experts either define case based legal vocabularies or use a pre-agreed legal vocabulary stored in a central public/privately-shared repositories (such as OntoMaven [9, 10], GitHub) and build legal rules based on it as shown before.

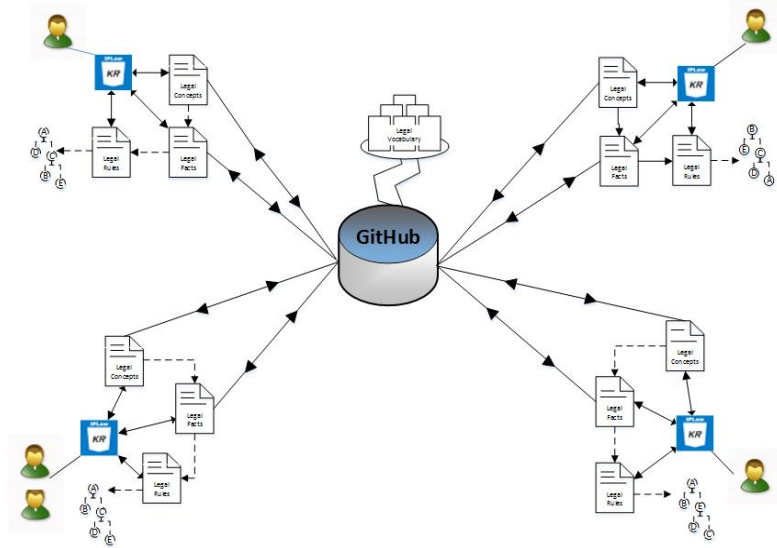

**Fig. 3.** Overview of 'KR4IPLaw' from a legal practitioner's perspective.

For storing thus generated rules in a machine oriented format and for interchanging such rules in a platform independent way, we translate them into XML using the language family of RuleML [11] as expression language on the PIM level. In particular, we make use of two complementary OASIS standards - OASIS Legal Document Markup Language, 'LegalDocML'[12] and OASIS Legal Rule Markup Language 'LegalRuleML',[13, 14] – for the XML-based legal-knowledge modeling and representation of legal norms and arguments. The SBVR vocabulary and facts are mapped on to an OWL2 ontology [15, 16] for generating the required knowledge base needed to provide the backend reasoning support for rule reasoner (e.g. prova [17]) embedded within a legal expert system. KR4IPLaw also provides an interface which allows legal practitioner/domain experts, who have little technical skills in the field of Semantic Web, to graphically inspect, modify and review ontology components and a second interface which provides knowledge modelers with tools for editing (with helpful features like syntax highlighting, autocompletion, and error-detection), validation, dependency-check, etc… for thus generated ontologies. Fig 4a and Fig 4b depicts these two interfaces.

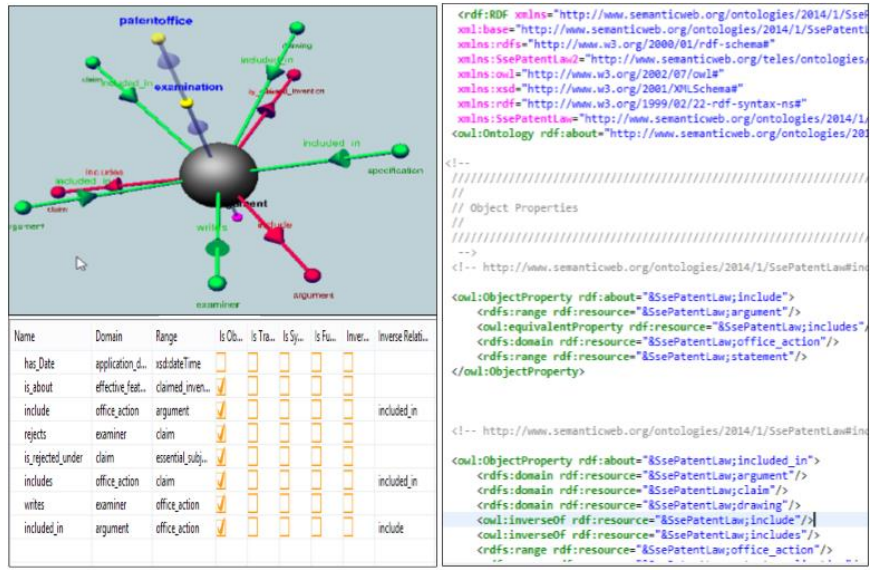

**Fig. 4(a).** Legal Practitioners/Domain Experts' perspective of KR4IPLaw (b) Knowledge Modeler's perspective of KR4IPLaw.

## 5   Conclusion

This paper presented how SBVR's Structured English (SSE) approach can be adapted to the legal domain. The proposed approach can act as a bridge between legal practitioner/domain experts and technical knowledge modelers. With an example from patent law we showed how SSE can help legal practitioners to build semi-formal procedural rules using legal vocabularies. As a proof-of-concept we introduced the tool, called KR4IPLaw. We further provided insights towards transforming these SSE computational independent legal rules to platform independent rules in OASIS LegalRuleML and W3C OWL, which thereafter can be translated into platform specific logical languages and reasoned using rule reasoners like Prova.

# References

1. Bézivin, J., Gerbé, O.: Towards a precise definition of the OMG/MDA framework. Automated Software Engineering, 2001.(ASE 2001). Proceedings. 16th Annual International Conference on. pp. 273–280 (2001).
2. Object Management.Group, Semantics of Business Vocabulary and Business Rules ( SBVR ). (2013).
3. Johnsen, Å., Berre, A.: A bridge between legislator and technologist - Formalization in SBVR for improved quality and understanding of legal rules. International Workshop on Business Models, Business Rules and Ontologies., Bressanone/Brixen, Italy (2010).
4. Johnsen, Å.: Semantisk modellering av juridisk regelverk med bruk av SBVR - en brobygger mellom jus og IT, (2011).
5. Schindler, S., Paschke, A., Ramakrishna, S.: Formal Legal Reasoning that an Invention Satisfies, SPL. AIIP2. , Bologna (2013).
6. Tommasi, M., Corallo, A.: SBEAVER: A Tool for Modeling Business Vocabularies and Business Rules. In: Gabrys, B., Howlett, R., and Jain, L. (eds.) Knowledge-Based Intelligent Information and Engineering Systems SE - 137. pp. 1083–1091. Springer Berlin Heidelberg (2006).
7. Alessio Bosca, Dario Bonino, P.P.: Ontosphere: more than a 3d ontology visualization tool.
8. Title 35 of the United States Code. (1952).
9. Paschke, A.: OntoMaven: Maven-based Ontology Development and Management of Distributed Ontology Repositories. CoRR. abs/1309.7341, (2013).
10. Paschke, A.: OntoMaven API4KB - A Maven-based API for Knowledge Bases. SWAT4LS (2013).
11. Boley, H., Paschke, A., Shafiq, O.: RuleML 1 . 0 : The Overarching Specification of Web Rules. RuleML. pp. 162–178 (2010).
12. Thomas F. Gordon: The Legal Knowledge Interchange Format ( LKIF ). (2008).
13. Palmirani, M., Governatori, G., Rotolo, A., Tabet, S., Boley, H., Paschke, A.: LegalRuleML: XML-Based Rules and Norms. In: Olken, F., Palmirani, M., and Sottara, D. (eds.) Rule - Based Modeling and Computing on the Semantic Web. pp. 298–312. Springer Berlin Heidelberg (2011).
14. Paschke, A., Ramakrishna, S.: Legal RuleML Tutorial Use Case - LegalRuleML for Legal Reasoning in Patent Law, (2013).
15. Karpovic, J., Nemuraite, L.: Transforming SBVR Business Semantics into Web Ontology Language OWL2 : Main Concepts. 231–254.
16. Reynares, E., Caliusco, M.L., Galli, M.R.: Approaching the feasibility of SBVR as modeling language for ontology development: An exploratory experiment. Expert Syst. Appl. 41, 1576–1583 (2014).
17. Kozlenkov, A.: Prova Rule Language Version 3.0 User's Guide. Internet http//prova. ws/index. html. (2010).